# Sarcasm Detection in Tweets with BERT and GloVe Embeddings

**Akshay Khatri, Pranav P and Dr. Anand Kumar M**
Department of Information Technology
National Institute of Technology
Karnataka, Surathkal
`{akshaykhatri0011, hsr.pranav}@gmail.com`
`m_anandkumar@nitk.edu.in`

## Abstract

Sarcasm is a form of communication in which the person states opposite of what he actually means. It is ambiguous in nature. In this paper, we propose using machine learning techniques with BERT and GloVe embeddings to detect sarcasm in tweets. The dataset is preprocessed before extracting the embeddings. The proposed model also uses the context in which the user is reacting to along with his actual response.

## 1 Introduction

Sarcasm is defined as a sharp, bitter or cutting expression or remark and is sometimes ironic (Gibbs et al., 1994). To identify if a sentence is sarcastic, it requires analyzing the speaker's intentions. Different kinds of sarcasm exist like propositional, embedded, like-prefixed and illocutionary (Camp, 2012). Among these, propositional requires the use of context.

The most common formulation of sarcasm detection is a classification task (Joshi et al., 2017). Our task is to determine whether a given sentence is sarcastic or not. Sarcasm detection approaches are broadly classified into three types (Joshi et al., 2017) . They are: Rule based, deep learning based and statistical based. Rule based detectors are simple, they just look for negative response in a positive context and vice versa. It can be done using sentiment analysis. Deep learning based approaches use deep learning to extract features and the extracted features are fed into a classifier to get the result. Statistical approach use features related to the text like unigrams, bigrams etc and are fed to SVM classifier.

In this paper, we use BERT embeddings (Devlin et al., 2018) and GloVe embeddings (Pennington et al., 2014) as features. They are used for getting vector representation of words. These embeddings are trained with a machine learning algorithm. Before extracting the embeddings, the dataset also needs to be processed to enhance the quality of the data supplied to the model.

## 2 Literature Review

There have been many methods for sarcasm detection. We discuss some of them in this section. Under rule based approaches, Maynard and Greenwood (2014) use hashtag sentiment to identify sarcasm. The disagreement of the sentiment expressed by the hashtag with the rest of the tweet is a clear indication of sarcasm. Vaele and Hao (2010) identify sarcasm in similes using Google searches to determine how likely a simile is. Riloff et al. (2013) look for a positive verb and a negative situation phrase in a sentence to detect sarcasm.

In statistical sarcasm detection, we use features related to the text to be classified. Most approaches use bag-of-words as features (Joshi et al., 2017). Some other features used in other papers include sarcastic patterns and punctuations (Tsur et al., 2010), user mentions, emoticons, unigrams, sentiment-lexicon-based features (González-Ibáñez et al., 2011), ambiguity-based, semantic relatedness (Reyes et al., 2012), N-grams, emotion marks, intensifiers (Liebrecht et al., 2013), unigrams (Joshi et al., 2015), bigrams (Liebrecht et al., 2013), word shape, pointedness (Ptáček et al., 2014), etc. Most work in statistical sarcasm detection relies on different forms of Support Vector Machines(SVMs) (Kreuz and Caucci, 2007). (Reyes et al., 2012) uses Naive Bayes and Decision Trees for multiple pairs of labels among irony, humor, politics and education. For conversational data,sequence labeling algorithms perform better than classification algorithms (Joshi et al., 2016). They use SVM-HMM and SEARN as the sequence labeling algorithms (Joshi et al., 2016).

For a long time, NLP was mainly based on statistical analysis, but machine learning algorithms have now taken over this domain of research providing unbeaten results. Dr. Pushpak Bhattacharyya, a well-known researcher in this field, refers to this as "NLP-ML marriage". Some approaches use similarity between word embeddings as features for sarcasm detection. They augment these word embedding-based features with features from their prior works. The inclusion of past features is key because they observe that using the new features alone does not suffice for an excellent performance. Some of the approaches show a considerable boost in results while using deep learning algorithms over the standard classifiers. Ghosh and Veale (2016) use a combination of CNN, RNN, and a deep neural network. Another approach uses a combination of deep learning and classifiers. It uses deep learning(CNN) to extract features and the extracted features are fed into the SVM classifier to detect sarcasm.

## 3 Dataset

We used the twitter dataset provided by the hosts of shared task on Sarcasm Detection. Initial analysis reveals that this is a perfectly balanced dataset having 5000 entries. There are an equal number of sarcastic and non-sarcastic entries in it. It includes the fields label, response and context. The label specifies whether the entry is sarcastic or non-sarcastic, response is the statement over which sarcasm needs to be detected and context is a list of statements which specify the context of the particular response. The test dataset has 1800 entries with fields ID(an identifier), context and response.

Most of the time, raw data is not complete and it cannot be sent for processing(applying models). Here, preprocessing the dataset makes it suitable to apply analysis on. This is an extremely important phase as the final results are completely dependent on the quality of the data supplied to the model. However great the implementation or design of the model is, the dataset is going to be the distinguishing factor between obtaining excellent results or not. Steps followed during the preprocessing phase are: (Mayo)

- Check out for null values - Presence of null values in the dataset leads to inaccurate predictions. There are two approaches to handle this:
    - Delete that particular row - We will be using this method to handle null values.
    - Replace the null value with the mean, mode or median value of that column - This approach cannot be used as our dataset contains only text.

- Tokenization and remove punctuation - Process of splitting the sentences into words and also remove the punctuation in the sentences as they are of no importance for the given specific task.

- Case conversion - Converting the case of the dataset to lowercase unless the case of the whole word is in uppercase.

- Stopword removal - These words are a set of commonly used words in a language. Some common English stop words are "a", "the", "is", "are" and etc. The main idea behind this procedure is to remove low value information so as to focus on the important information.

- Normalization - This is the process of transforming the text into its standard form. For example, the word "gooood" and "gud" can be transformed to "good", "b4" can be transformed to "before", ":)" can be transformed to "smile" its canonical form.

- Noise removal - Removal of all pieces of text that might interfere with our text analysis phase. This is a highly domain dependent task. For the twitter dataset noise can be all special characters except hashtag.

- Stemming - This is the process of converting the words to their root form for easy processing.

Both training and test data are preprocessed with the above methods. Once the above preprocessing steps have been applied we are ready to move to the model development.

## 4 Methodology

In this section, we describe the methods we used to build the model for the sarcasm detection.

### 4.1 Feature Extraction

Feature extraction is an extremely important factor along with pre-processing in the model building process. The field of natural language processing(NLP), sentence and word embeddings are majorly used to represent the features of the language. Word embedding is the collective name for a set of feature learning techniques in natural language processing where words or phrases from the vocabulary are mapped to vectors of real numbers. In our research, we used two types of embeddings for the feature extraction phase. One being **BERT(Bidirectional Encoder Representations from Transformers)** word embeddings (Devlin et al., 2018) and the other being **GloVe(Global Vectors) embeddings** (Pennington et al., 2014).

### 4.1.1 BERT embeddings

'Bert-as-service' (Xiao, 2018) is a useful tool for the generation of the word embeddings. Each word is represented as a vector of size 768. The embeddings given by BERT are contextual. Every sentence is represented as a list of word embeddings. The given training and test data has response and context as two fields. Embeddings for both context and response were generated. Then, the embeddings were combined in such a way that context comes before response. The intuition to this being that it is the context that elicits a response from a user. Once the embeddings are extracted, the sequence of the embeddings were padded to get them to the same size.

### 4.1.2 GloVe embeddings

The results given by BERT not being up to the mark led us to search for a twitter specific embedding and thus we chose GloVe embeddings specifically trained for twitter. It uses unsupervised learning for obtaining vector representation of words. The embeddings given by GloVe are non-contextual. Here we decided to choose GloVe twitter sentence embeddings for training the models as it would capture the overall meaning of the sentence in a relatively lesser amount of memory. This generated a list of size 200 for each input provided. Once the sentence embeddings were extracted, the context and the response were combined such that context comes before response. Context embeddings were generated independent of the response so that the sentiment of response would not effect the sentiment of the context.

## 4.2 Training and Predictions

After extraction of the word embeddings, the next step is to train these to build a model which can be used to predict the class of test samples. Classifiers like Linear Support Vector Classifier(LSVC), Logistic Regression(LR), Gaussian Naive Bayes and Random Forest were used. Scikit-learn (Pedregosa et al., 2011) was used for training these models. Word embeddings were obtained for the test dataset in the same way mentioned before. Now, they are ready for predictions.

## 5 Reproducibility

### 5.1 Experimental Setup

Google colab with 25GB RAM was used for the experiment, which includes extraction of embedding, training the models and prediction.

### 5.2 Extracting BERT embeddings

We use the bert as a service for generating the BERT embeddings. We spin up BERT as a service server and create a client to get the embeddings. We use the uncased_L-12_H-768_A-12 pretrained BERT model to generate the embeddings.

All of the context(i.e 100%) provided in the dataset was used for this study. Word embeddings for the response and context are generated separately. Embeddings for each word in the response is extracted separately and appended to form a list. Every sentence in the context is appended one after the other. The same is done for response embeddings. The embeddings of the context and that of response fields are concatenated to get the final embedding.

### 5.3 Extracting the GLOVE embeddings

We used genism to download glove-twitter-200 pretrained model. Embeddings for the response and context are extracted separately. Sentences in the given context are appended to form a single sentence. Later we generate the sentence embeddings for response and context separately. The context embeddings and response embeddings are concatenated to generate the final embedding.

### 5.4 Training the model

We use the Scikit-learn machine learning library to train the classifiers(SVM, Logistic Regression, Gaussian Naive Bayes and Random Forest). Trained models are saved for later prediction. Using the saved models, we predict the test samples as SARCASM or NOT_SARCASM.

## 6 Results

The result was measured using the metric F-measure. F-measure is a measure of a test's ac-curacy and is calculated as the weighted harmonic mean of the precision and recall of the test (Zhang and Zhang, 2009). Now, we discuss the results obtained with BERT and GloVe separately.

| Classifier | F-measure |
|---|---|
| Linear Support Vector Classifier | 0.598 |
| Logistic Regression | 0.6224 |
| Gaussian Naive Bayes | 0.373 |
| Random Forest | 0.577 |

Table 1: Results with BERT embeddings excluding context

| Classifier | F-measure |
|---|---|
| Linear Support Vector Classifier | 0.616 |
| Logistic Regression | 0.630 |

Table 2: Results with BERT embeddings including both context and response

### 6.1 With BERT

Among the classifiers mentioned in the previous section, a good result was received from SVM and logistic regression, with the latter giving the best results. Table 1 shows the results of training the classifiers only on the response and excluding the context. Table 2 shows the results obtained with BERT including the context. It is clear that taking the context into consideration boosts the result.

### 6.2 With GloVe

The results for this approach gave much better results when compared to the BERT embedding approach. Also, GloVe was much faster than BERT. Among the two classifiers logistic regression gave the better results of the two. Table 3 shows the results obtained with GloVe.

## 7 Conclusion

Sarcasm detection can be done effectively using word embeddings. They are extremely useful as they capture the meaning of a word in a vector representation. Even though BERT gives contextual word representations i.e the same word occuring multiple times in a sentence may have different vectors, it didn't perform to the mark when compared to GloVe which gives the same vector for a word occuring multiple times. However, this cannot be generalized. It may depend on the dataset. Among the classifiers, logistic regression always outperformed the other classifiers used in this study.

| Classifier | F-measure |
|---|---|
| Linear Support Vector Classifier | 0.679 |
| Logistic Regression | 0.690 |

Table 3: Results with GloVe embeddings including both context and response

## Acknowledgments

We want to thank all the reviewers for their useful suggestions to improvise the paper. These recommendations have aided us in getting this paper into the form it is now. We want to thank the organizers of the shared task for giving us the opportunity to work in this exciting field.